\documentclass{article}

\usepackage[nonatbib]{neurips_2023}
\usepackage[utf8]{inputenc} 
\usepackage[T1]{fontenc}    
\usepackage{hyperref}       
\usepackage{url}            
\usepackage{booktabs}       
\usepackage{amsfonts}       
\usepackage{nicefrac}       
\usepackage{microtype}      
\usepackage{xcolor}         
\usepackage{amsmath}
\usepackage{multirow}
\usepackage{graphicx}
\pdfoutput=1
\graphicspath{ {./images/} }

\title{GAT-GAN : A Graph-Attention-based Time-Series Generative Adversarial Network}

\author{%
  Srikrishna Iyer$^{1}$ \quad Teng Teck Hou$^{2}$ \\
  Data-Analytics Strategic Technology Centre\\
  Group Technology Office, ST Engineering Pte Ltd, Singapore\\  \texttt{\{srikrishna.rameshiyer$^{1}$,thteng$^{2}$\}@stengg.com} \\
}

\begin{document}
\nolinenumbers
\maketitle

\begin{abstract}
  Generative Adversarial Networks (GANs) have proven to be a powerful tool for generating realistic synthetic data. However, traditional GANs often struggle to capture complex relationships between features which results in generation of unrealistic multivariate time-series data. In this paper, we propose a Graph-Attention-based Generative Adversarial Network (GAT-GAN) that explicitly includes two graph-attention layers, one that learns temporal dependencies while the other captures spatial relationships. Unlike RNN-based GANs that struggle with modeling long sequences of data points, GAT-GAN generates long time-series data of high fidelity using an adversarially trained autoencoder architecture. Our empirical evaluations, using a variety of real-time-series datasets, show that our framework consistently outperforms state-of-the-art benchmarks based on \emph{Frechet Transformer distance} and \emph{Predictive score}, that characterizes (\emph{Fidelity, Diversity}) and \emph{predictive performance} respectively. Moreover, we introduce a Frechet Inception distance-like (FID) metric for time-series data called Frechet Transformer distance (FTD) score (lower is better), to evaluate the quality and variety of generated data. We also found that low FTD scores correspond to the best-performing downstream predictive experiments. Hence, FTD scores can be used as a standardized metric to evaluate synthetic time-series data. 
\end{abstract}

\section{Introduction}

Extensive work has been done in capturing temporal dependencies for generating time-series data \cite{ mogren_c-rnn-gan_2016,esteban_real-valued_2017,yu_seqgan_2017,yoon_time-series_2019,wang_continuous_2019,ni_conditional_2020,wiese_quant_2020}. However, for a myriad of applications including but not limited to transportation \cite{shao_traveling_2018}, social science \cite{kupilik_spatio-temporal_2018}, criminology \cite{rumi_modelling_2020} and non-linear systems \cite{rydin_gorjao_open_2020}, extracting dynamically changing spatial dependencies poses a unique challenge for generative modeling. The generative model should not only encode complex temporal dynamics of multivariate distributions but also extract spatial relationships between feature  distributions. For example, a non-linear system can have multiple correlating factors influencing the predicted output which are dynamically changing at each time step \cite{altosole_real-time_2009}. An additional problem is that RNN-based (Recurrent neural networks) architectures fail to extract long-range dependencies due to the vanishing gradient problem and get easily influenced by sequences closer to the current time step \cite{zhao_rnn_2020}. Graph-based models  have been introduced to extract spatiotemporal embeddings in combination with attention to determine node importance without prior knowledge of the graph structure \cite{velickovic_graph_2018}. Thus, it is of utmost importance to design a task-independent and generalized synthetic data generation technique using GANs for longer sequences, while preserving spatial correlations. 

In this work, we make three main contributions: i) GAT-GAN, a graph-attention-based GAN framework for time series generation that preserves both temporal and spatial dynamics. We propose an adversarially trained autoencoder, where the encoder (generator) maps the posterior distribution of the latent space variables to a priori distribution, while the decoder maps it back to the original distribution. The model is jointly trained using two losses namely, a reconstruction loss arising from the autoencoder and an adversarial loss from real and synthetic samples. Two separate spatial and temporal graph attention layers are used to train GAT-GAN, thereby explicitly encouraging the model to extract dynamic correlations between variables and temporal dependencies respectively. As far as we know, our approach is the first to leverage graph-attention networks in an adversarially trained autoregressive GAN framework ii) We  demonstrate that samples generated by GAT-GAN are of sufficient fidelity and diversity for short and long sequence lengths using real-world time-series datasets. Based on our experiments, we found that GAT-GAN achieves significant improvements over the state-of-the-art across multiple sequence lengths, in generating realistic time series. Additionally, we use the predictive score (Train on synthetic, test on real) to evaluate its predictive performance on a downstream forecasting task  iii) We propose a unified metric called Frechet Transformer distance (FTD) to evaluate the fidelity and diversity of generated samples by utilizing latent representations of a pre-trained Transformer \cite{zerveas_transformer-based_2021}. We prove that the lowest FTD values correspond to the best-performing models in the downstream forecasting task measured using a predictive score. Thus, the FTD score can be used to potentially benchmark GAN models for various applications like forecasting. 

We structure this work as follows: We review the related work in section 2 and formulate the objectives of this paper in section 3. In section 4, we propose the GAT-GAN model followed by section 5, where we evaluate the GAT-GAN model with the benchmarks based on our proposed FTD score and a downstream forecasting task quantified by predictive score. Additionally, we also perform an ablation study to demonstrate the importance of every component in the model. Finally, in section 6, we conclude this work.
 
\section{Related Work}

GANs have been an active area of research that has been predominantly used to generate images and language. They have only recently garnered attention in the time-series domain \cite{mogren_c-rnn-gan_2016,esteban_real-valued_2017,yu_seqgan_2017,yoon_time-series_2019,wang_continuous_2019,ni_conditional_2020,wiese_quant_2020} for data generation, forecasting \cite{wu_adversarial_2020} and augmentation \cite{paul_psa-gan_2022}. Most of the papers have adopted the vanilla GAN architecture using recurrent networks (RNNs and LSTMs) or convolution networks (CNNs), where data is generated sequentially to model temporal dynamics. The RNN-based framework has been applied to generate synthetic data in numerous domains such as finance \cite{wiese_quant_2020}, music \cite{dong_musegan_2018,engel_gansynth_2018}, bio-signals \cite{ni_conditional_2020} and electrical systems \cite{wang_generating_2020}. However, RNN-based models have typically failed to capture long-range and spatial dependencies, attributed to their inherent sequential nature which prevents parallelization during training. Convolution networks (CNNs) can extract spatial information using multiple kernels and have been used in combination with RNNs \cite{zuo_convolutional_2015,shi_convolutional_2015}. However, the underlying sequential computation constraint still remains an issue. Recently, Transformers \cite{vaswani_attention_2017} have been adversarially trained to generate synthetic multivariate data \cite{li_tts-gan_2022}. Though the authors were successful in extracting global dependencies, transformers are theoretically incapable of modeling periodic or recursion properties in sequences on their own, either with soft or hard attention for long sampled sequences \cite{hahn_theoretical_2020}. Practically, this can be circumvented using a large number of layers and attention heads, thereby quadratically increasing computational complexity $\mathcal{O}(n^2)$, where $n$ is the length of sequence \cite{wang_linformer_2020}. However, self-attention used in conjunction with convolutional feature maps has been successful in addressing two fundamental issues - better generalization at faster converging speeds due to convolutional layers and a larger receptive field (more contextual information) contributed by self-attention networks \cite{dai_coatnet_2021}. Hence, this work utilizes 1D convolution layers in the encoder (generator) in combination with self-attention applied to a connected acyclic undirected graph.

Another line of work in the multivariate time-series domain is focused on representing spatial interactions between variable distributions. Graphs have emerged as a generalized feature representation method for both structured and unstructured data. Unlike images and text-based representations, graphs are capable of representing data without a fixed form, that does not conform to a grid-like structure. Thus, graph-based generative models can be applied to four major spatiotemporal datatypes that include Spatiotemporal events (eg. crime detection), trajectory data (eg. traffic flow), spatiotemporal graphs (eg. social networks), and multivariate time-series data (eg. IoT systems) \cite{gao_generative_2022} . Graph neural networks (GNNs) used recurrent neural networks to process cyclic directed and undirected graphs \cite{gori_new_2005,Scarselli_GNN_2009}. This idea was improved upon by directly applying convolutions to the graph structures. One of the main challenges to this approach is to define a convolution operator that can be applied to variable-sized neighborhoods. Hence, it requires a separate weight matrix to define the neighborhood of each input channel leading to increasing computational complexity. While there have been several models to normalize neighborhoods to a fixed number of nodes \cite{duvenaud_convolutional_2015,atwood_diffusion-convolutional_2016,hamilton_inductive_2017}, they were outperformed by an attention-based architecture that computes latent representations of each node in the graph by attending over its neighbors using self-attention \cite{velickovic_graph_2018}. However, \cite{brody2022how} argue that the graph attention network does not compute dynamic attention i.e. the network is unable to take into account the specific node that is being attended to when determining the importance of its neighbors. Hence, by modifying the order of internal operations in the attention function, the proposed work in \cite{brody2022how} was able to implement a more expressive form of attention.

Inspired by the recent work in graph attention \cite{brody2022how} and convolutional networks, we propose a graph-attention network \cite{brody2022how} based architecture to generate time-series data while preserving spatial information. Firstly, we design a spatial attention component, where each feature is represented by a node, to capture spatial dependencies and relationships between the variables. Additionally, we encode temporal dependencies using attention, where each time step is represented by a node, to access distant features without any neighborhood restrictions.  

Finally, the selection of a suitable metric to evaluate generated time-series samples is another challenge. Frechet Inception distance (FID) \cite{salimans_improved_2016} and Inception score \cite{heusel_gans_2017} are standardized scores widely used to evaluate fidelity and diversity in the computer vision domain. Several works have also improved upon the FID and inception score to include class information \cite{liu_improving_2023}, speed up computation time \cite{mathiasen_backpropagating_2021}, include spatial features \cite{nash_generating_2021}, and eliminate bias training examples \cite{chong_effectively_2020}. However, these metrics have only been used to evaluate the quality of synthetic image-generating models. The authors in \cite{barua_quality_2019} propose a metric called cross-local intrinsic dimensionality (crossLID) to measure the degree of closeness of distributions. Though they can be used for time series, it is unclear whether the score can be applied to high-dimensional data where the variance cannot be captured by a local dimensionality. Perceptual path length (PPL) \cite{karras_style-based_2021} is another metric introduced recently to measure the perceptual difference (calculated using spherical interpolation) between two images in the latent space. However, the embeddings are obtained using VGG16, which is computationally expensive to use as a loss function during model hyperparameter tuning. Hence, we propose an FID-based score called  Frechet transformer distance (FTD) for synthetic time-series data that uses transformer-based learning representations \cite{zerveas_transformer-based_2021} to compute the Frechet distance between synthetic and real embeddings. Additionally, building on the \emph{Train on synthetic – Test on real (TSTR)} setup used to score the quality of synthetic samples in downstream classification/prediction tasks \cite{esteban_real-valued_2017,yoon_time-series_2019}, we evaluate the performance of GAN models on a downstream forecasting task of a fixed prediction window. We also show empirically that the GANs with the lowest FTD scores correspond to the best-performing forecasting model. 

\section{Problem formulation}

Consider a multi-variate time-series dataset that consists of temporal features $x_t=[x_{t-\tau+1},..,x_t]$ (values that change across time) and spatial features $x_t^f = [x_t^1,x_t^2,...x_t^F]$ (attributes across a single time step, e.g. Voltage, current). Let $x_t^{f,j} \in \Re$ denote the value of the $f^{th}$ feature at index $j$ at time $t$, then $x_t^j \in \Re^\tau$ are the values in a sequence $\tau$ of index $j$ at time $t$. Also, $X_t = [x_t^1,x_t^2,..,x_t^K]^T \in \Re^{K \times \tau}$ denote the values of all sequences at all indices at time $t$. So, the training data is denoted by $\chi = [X_1,X_2,...,X_f]^T \in \Re^{K \times \tau \times F}$ are the values of all the features of all the indices across $\tau$ sequences with a joint distribution $P(\chi)$. 

Our main objective is to model a $\bar{P}(\chi)$ that best approximates $P(\chi)$. Instead of doing so directly using an adversarial framework, we introduce an autoregressive function with deep encoding and decoding functions. Let $Z \in \Re^{\bar{K} \times \bar{\tau} \times \bar{F}}$ be the latent code vector of an autoencoder and $P(Z)$ be the prior distribution of the latent space. Then, the conditional encoding distribution is given by $Q(Z|\chi)$ which is aggregated to define the posterior distribution of $Q(Z)$ as : 
\begin{equation} \label{eq1}
Q(Z) = \sum_\chi Q(Z|\chi) P(\chi)
\end{equation}

The joint posterior $Q(Z)$ is matched to the prior $P(Z)$ using an adversarial framework, while the autoencoder minimizes the reconstruction error to match the encoded conditional distribution $Q(Z|\chi)$ to its decoded distribution $P(Z|\chi)$. The encoder generates the aggregated posterior distribution to outsmart the discriminator into thinking that $Q(Z)$ is sampled from the real prior distribution $P(Z)$. The autoencoder and adversarial network are trained in batches using stochastic gradient descent in two phases : 
\begin{equation} \label{eq2}
Reconstruction \, phase : min \, L[Q(Z|\chi),P(Z|\chi)]
\end{equation}
\begin{equation} \label{eq3}
Adversarial \, phase : min \, L[Q(Z),P(Z)]
\end{equation}
where L is a loss function between the distributions.
In the reconstruction phase, the reconstruction error between the encoder and decoder is minimized. In the adversarial phase, the discriminator distinguishes the generated latent joint distribution from the true prior. Finally, the error is used to update the generator (encoder) to fool the discriminator. 

The proposed model differs from a state-of-the-art autoregressive-based model \cite{yoon_time-series_2019} by eliminating the use of a separate generator. Instead, the authors in \cite{makhzani_adversarial_2016} posit an elegant solution to train the encoder network as the universal approximator of the posterior. Consider an encoder function $\epsilon(\chi,\phi)$ that takes an additional input, noise $\phi$ (eg. Gaussian noise of fixed variance and mean) with a distribution $P(\chi_{\phi})$. The posterior distribution $Q(Z|\chi)$ and its aggregated posterior $Q(Z)$ can be reformulated as:
\begin{equation} \label{eq4}
Q(Z|\chi) = \sum_{\phi}Q(Z|\chi,\phi) P(\chi_{\phi}) \Longrightarrow Q(Z) = \sum_\chi \sum_\phi Q(Z|\chi,\phi) P(\chi)P(\chi_{\phi})
\end{equation}
Here, we assume that $Q(Z|\chi)$ is a deterministic function of $\chi$ and random noise $\phi$. Unlike a vanilla autoencoder, the two sources of stochasticity (data distribution and random noise) eliminate gaussian constraints within the posterior $Q(Z|\chi)$. Hence, through adversarial training, the encoder $\epsilon(\chi,\phi)$ can generalize over any posterior distribution given an input $\chi$ to match $Q(Z)$ with the real prior $P(Z)$. 

\section{Proposed Model: Graph Attention Generative Adversarial Networks (GAT-GAN)}

\begin{figure}[h]
\centering
\includegraphics[width=\textwidth]{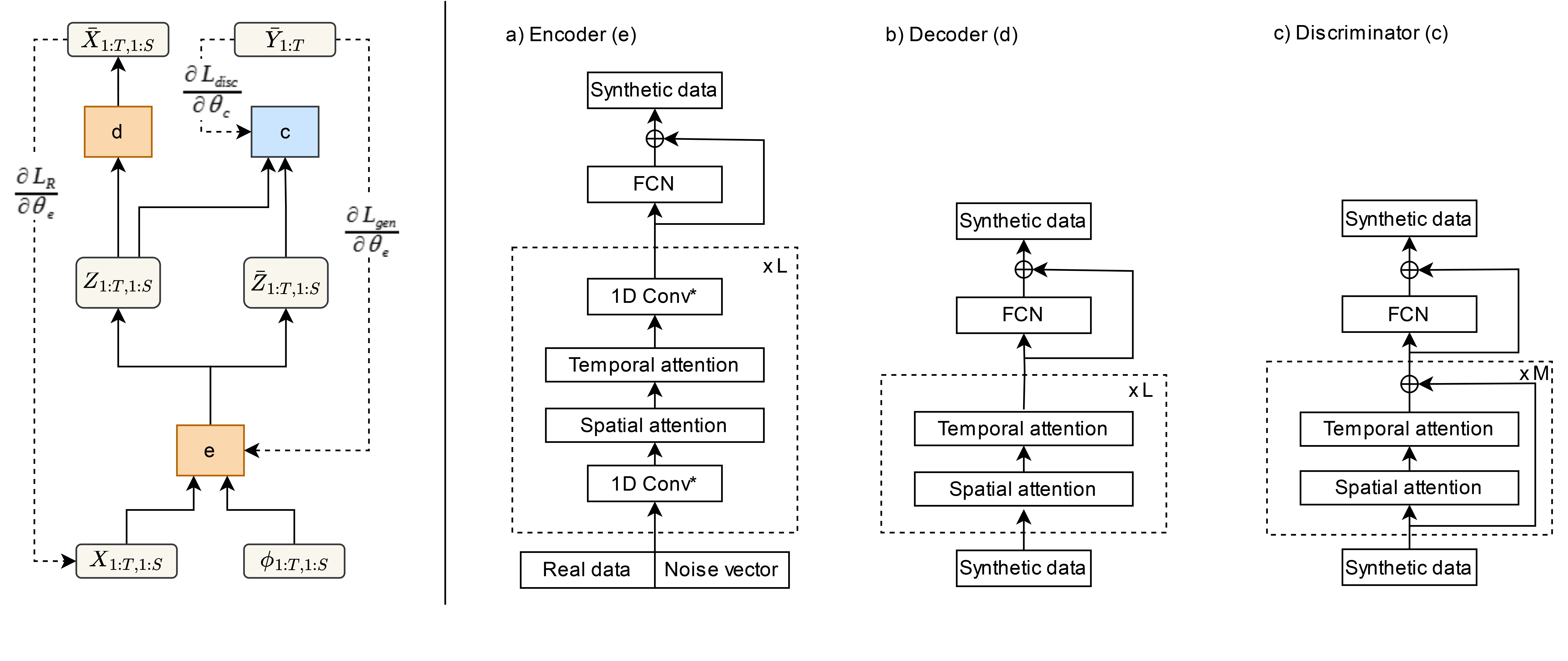}
\caption{\emph{Left}: Block diagram of the proposed GAT-GAN model, where the dotted lines represent back-propagation of gradients and solid lines indicate forward-propagation of data.  \emph{Right}: Block diagrams of the components: Encoder, Decoder, and Discriminator. a) The encoder consists of \emph{N} number of spatial and temporal attention layers with two spectral normalized convolution layers and M full connected neural network layers, where \emph{M} and \emph{N} are hyperparameters. b) The decoder and c) discriminator have similar architectures with the only difference in discriminator is residual connections in \emph{N} temporal and spatial attention networks to converge faster.}
\label{fig1}
\end{figure}

Figure \ref{fig1} outlines the overall framework of the proposed GAT-GAN model in this paper. It consists of three components: An encoding function which also serves as the generator, a decoding function, and a discriminator. The autoregressive components are trained together with the discriminator with dual objectives - a reconstruction error and an adversarial training criterion. The three components share the same network structure, where each of them consists of several spatial-temporal graph attention blocks followed by a fully-connected layer. To optimize training efficiency, we use residual connections in all three components \cite{he_deep_2016} and employ spectral normalization in 1D convolution layers \cite{miyato_spectral_2018} used solely in the encoder to stabilize training of the encoder (generator). 

\subsection{Encoder}

The encoder (Generator) module maps the input temporal and spatial features to a lower-dimensional latent vector representation. Let $Z \in \Re^{\hat{K} \times \hat{\tau} \times \hat{F}}$ be the latent representation of the input feature $\chi$, then the encoder function that maps the input features to their latent codes is given by : 
\begin{equation} \label{eq5}
\begin{split}
e_{(p-1)} = \Re^{K \times \tau \times F} \longrightarrow \Re^{\hat{K} \times \hat{\tau} \times \hat{F}} \\
\chi^{(l-1)} \longrightarrow Z^{(l-1)} = \epsilon(\chi)
\end{split}
\end{equation}
where $\epsilon$ is a set of trainable blocks ${\epsilon^p}$ with $p=[0,1,2,3]$, implemented as a function that comprises four trainable blocks that include a 1D convolution layer, Spatial graph attention, Temporal graph attention, and fully connection neural network. Before the input data is passed into the graph attention layers, a standard convolution layer is applied in the temporal dimension to extract global dependencies for each feature : 
\begin{equation} \label{eq6}
\chi^{(l-1)} = LeakyReLU(\hat{\sigma}(\theta \otimes \chi^{(l-1)})) \in \Re^{K \times \tau_{(l-1)} \times F_{(l-1)}}
\end{equation}
where $\chi^{(l-1)} = [X_1,X_2,...,X_{f_{(l-1)}}]^T \in \Re^{K \times \tau_{(l-1)} \times F_{(l-1)}}$ with $K$ number of sequences, $\tau_{(l-1)}$ is the sequence length and $F_{(l-1)}$ is the number of features of the input data in the $l^{th}$ convolution layer. $\otimes$ denotes 1D convolution operation with $\theta$ tunable parameters of the convolution kernel. $\hat{\sigma}$ refers to spectral normalization \cite{miyato_spectral_2018} to improve stabilization of the convolution operation. The convolution feature map is then passed through a spatial graph attention mechanism to capture spatial correlations between the features. Here, we use the attention mechanism described in \cite{brody2022how} to adaptively capture dynamic relationships between nodes in the spatial domain. 
Thus, consider an undirected graph G = (n,e), where n is a set of $n = \{N\}$ nodes; e is a set of $e = \{E\}$ edges which signify the connectivity between the nodes. For a spatial-oriented graph, each node represents a feature and the edges denote the relationship between the features. Hence, spatial attention is given by :
\begin{equation} \label{eq7}
z^{(l-1)} = \sigma(\sum_{k\in N_j} \alpha_{jk}.W_1\chi_k^{(l-1)})
\end{equation}
Where the attention is given by :
\begin{equation} \label{eq8}
\alpha_{jk} = softmax_k(W_2^T LeakyReLU(W_1.[\chi_j^{(l-1)}||\chi_k^{(l-1}])+b_s)
\end{equation}
Here, $\chi^{(l-1)} \in \Re^{K \times \tau_{(l-1)} \times F_{(l-1)}}$ is input of the $l^{th}$ spatial attention block. $W_1 \in \Re^{2F^{(l-1)} \times 2F^{(l-1)}}$, $W_2 \in \Re^{2F^{(l-1)}}$, $b_s \in \Re^{F_{(l-1)} \times F_{(l-1)}}$ are tunable weights, weights of the neural network and bias respectively. $\sigma$ is the sigmoid activation function. The attention matrix $\alpha_{jk} \in \Re^{K \times F_{(l-1)} \times F_{(l-1)}}$ signifies the strength of relationship between feature-oriented nodes $j$ and $k$.
The resultant spatial attention embeddings are passed to the temporal graph attention layer. In the temporal domain, there exist relationships between each time step of a sequence across multiple features. Hence, for a time-oriented graph, each node represents a time step and edges denote the relationship between the time steps : 
\begin{equation} \label{eq9}
z^{(l-1)} = \sigma(\sum_{k\in N_j} \alpha_{jk}.V_1\chi_k^{(l-1)})
\end{equation}
Where the attention is given by :
\begin{equation} \label{eq10}
\alpha_{jk} = softmax_k(V_2^T LeakyReLU(V_1.[\chi_j^{(l-1)}||\chi_k^{(l-1}])+b_u)
\end{equation}
where, $V_1 \in \Re^{2F^{(l-1)} \times 2F^{(l-1)}}$ , $W_2 \in \Re^{2F^{(l-1)}}$, $b_s \in \Re^{\tau_{(l-1)} \times \tau_{(l-1)}}$ are tunable parameters. The attention matrix $\alpha_{jk} \in \Re^{K \times \tau_{(l-1)} \times \tau_{(l-1)}}$ signifies the strength of relationship between time-oriented nodes $j$ and $k$. To smoothen out the resultant output, a 1D convolution layer with spectral normalization and average pooling is applied. Finally, a feed-forward network is applied to create high-level features from the multivariate sequence $z^{(l-1)}$ to further extract embeddings $z^{(r-1)}$ where $r$ is the number of feed-forward layers. These layers consist of two sequential fully connected layers with LeakyReLU activation and Batch normalization. Additionally, a residual connection is used to allow the encoder to learn the attention-based hidden representations and converge faster.
\begin{equation} \label{eq11}
z^{(r-1)} = [LeakyReLU(\sum_j^K W_j z^{(l-1)} + b)] + z^{(l-1)}
\end{equation}

\subsection{Decoder}

The architecture is similar to that of the encoder $\epsilon$. The only difference is the exclusion of the spectral normalized 1D convolution layer. The Decoder module maps the lower-dimensional vector space back to its original feature space. If $Z$ is the latent vector, the reconstructed feature space $\bar{\chi} \in \Re^{\bar{K} \times \bar{\tau} \times \bar{F}} $ is given by :
\begin{equation} \label{eq12}
\begin{split}
d_{p-1} : \Re^{\hat{K} \times \hat{\tau} \times \hat{F}} \longrightarrow \Re^{\bar{K} \times \bar{\tau} \times \bar{F}} \\
          Z \longrightarrow \bar{\chi} = \epsilon_m^3(\epsilon_l^2(\epsilon_l^1(Z)))
\end{split}
\end{equation}
where, $p \in [1,2,3,4]$ are decoder components, $\epsilon_r^3$ refers to $r$ layers of FCN, $\epsilon_l^1$ and $\epsilon_l^2$ denote the $l$ layers of spatial and temporal attention layers respectively.  

\subsection{Discriminator}

The architecture of the discriminator function mirrors the architecture of the decoder. It maps the encoder's output $z^{(m-1)}$ from input time-series $\chi^{(m-1)}$ and random noise vector $\phi^{(m-1)}$ to a probability score $y^{*(m-1)}$ : 
\begin{equation} \label{eq13}
\begin{split}
c_{p-1} : \Re^{*\bar{K} \times \bar{\tau} \times \bar{F}} \longrightarrow \Re^{\bar{K}} \\
    Z^{*(l-1)} \longrightarrow y^{*(m-1)} = \sigma(\epsilon_r^3(\epsilon_l^2(\epsilon_l^1(Z))))
\end{split}
\end{equation}
where $Z^* \in \Re^*$ refers to either real or synthetic latent representations. Likewise, $y^*$ is the probability score for real $y$ or synthetic $\hat{y}$ data. $p \in [1,2,3,4]$ are discriminator components, $\sigma$ is the Sigmoid activation function which converts the input sequence into a score.
 
\subsection{Loss functions}

Our initial objective is to ensure that the encoding and decoding functions accurately reconstruct the original data $\bar{\chi}$ of their original input $\chi$ from their latent representations $Z$. Hence, minimizing the reconstruction error as shown below allows a mapping between the feature and latent spaces : 
\begin{equation} \label{eq14}
L_r = E_{\chi}[\sum_j^K||\chi_j - \bar{\chi}_j||_{2}]
\end{equation}
During training, the generator receives both the synthetic embeddings of Gaussian noise $\phi$ and real data $\chi$ to generate a synthetic vector $Z^*$. The generator loss is calculated as a minimization of the likelihood of correctly classified synthetic embeddings and the reconstruction error,
\begin{equation} \label{eq15}
L_{gen} = E_{\phi}[\sum_j^K \log(y_j) + L_r + \rho]
\end{equation}
where $y$ refers to the classification score for real latent codes, $L_r$ is the reconstruction loss, $\rho$ is an error term used to stochastically flip the classification probabilities during training to prevent the gradients from vanishing and $K$ is the number of sequences. The discriminator loss is used to adversarially train the encoder (generator) to capture the conditional distributions in the data. As a result, the loss function is computed using the sum of classification probabilities of real and synthetic embeddings,
\begin{equation} \label{eq16}
L_{disc} = E_\chi[\sum_j^K \log(\hat{y}_j) + \log(1-y_j) + \rho]
\end{equation}
where $\bar{y}$ classification results for synthetic latent codes.

\section{Experiments}

Evaluating synthetic time series data generated by GAN models poses a challenge since most existing metrics are domain (eg. image data) and model dependent \cite{alaa_how_2022}. Hence we assess the GAN models based on two criteria : (i) Fidelity and Diversity: The quality of samples generated by measuring how indistinguishable it is from the real data and to what extent can the generated samples encompass the entire range of variations observed in real samples. (ii) Predictive performance: Measuring the ability of the model to generalize over an unseen test dataset in a downstream forecasting task.

For fidelity and diversity, we propose the \emph{Frechet Transformer distance} score to measure the closeness of the real data distribution to the synthetic distributions. Consider $\chi_r$ and $\chi_s$ be the real and synthetic data and their corresponding distributions, $P_r$ and $P_s$ respectively. Let $Q_r = supp(P_r)$ and $Q_s = supp(P_s)$  where the gaussian-support refers to the minimum volume subset of $Q = supp(P)$ that approximates a probability mass of the Gaussian $\phi$ function \cite{scholkopf_estimating_2001} : 
\begin{equation} \label{eq17}
Q^\phi \overset{\Delta}{=} \underset{q \subset Q}{min} V(q),  s.t. P(q) = \phi
\end{equation}
where $V(q)$ is the volume of q and the Gaussian function is normalized as $\phi \in [0,1]$. Hence, gaussian-support can be referred to as dividing the distribution $P$ into inliers denoted by $Q^\phi$ and outliers $\bar{Q}^\phi$, where Q = $Q^\phi \cup \bar{Q}^\phi$
To solve the optimization problem in (\ref{eq17}), we train an embedding function $E$ to map $Q$ onto a latent multidimensional Gaussian space $\zeta$. As shown in \cite{heusel_gans_2017}, we only consider the first moments: mean and covariance and measure the Frechet distance between the two Gaussians (synthetic and real) which is given by \cite{dowson_frechet_1982}: 
\begin{equation} \label{eq18}
d^2(\zeta_r(m_1,m_2),\zeta_s(\bar{m}_1,\bar{m}_2)) = {||m_1 - \bar{m}_1||}_2^2 + Tr(m_2 + \bar{m}_2 - 2\sqrt{m_2*\bar{m}_2})
\end{equation}
where $m1,m2$ refer to mean and covariance respectively and Tr is the trace of the matrix.
We replace the Inception \cite{szegedy_rethinking_2016} with the Transformer encoder of \cite{zerveas_transformer-based_2021} as the embedding function $E$, which we train separately for each dataset as a supervised regression task. here, the input sequences consisted of all values until the penultimate time-step $[x_{t-\tau+1},..,x_{t-1}]$, while the corresponding label is the sample at the last time step $[x_{t}]$. The embeddings are then obtained for real and synthetic data separately and the Frechet distance from (\ref{eq18}) is computed. 

For (ii), we report the predictive score \cite{yoon_time-series_2019} by training a post-hoc forecasting model (2-layer LSTM) on synthetic data for a prediction size $p = 8$ samples and varying context lengths $c = \tau - p$, where $\tau$ is the sequence length. The model is then tested on the real dataset and the mean absolute error (MAE) is measured between the predicted and real samples: 
\begin{equation} \label{eq19}
MAE = \sum_{i=1}^K\sum_{j=1}^p |y_{pred} - y|
\end{equation}
where $y$ is the last p samples of the real sequences $[x_{t-p+1},..,x_{t}]$ and $y_{pred}$ is the predicted output of the post-hoc LSTM model trained on synthetic data $[x_1,..,x_{t-p}]$. 

Lastly, we conduct an ablation study of our proposed model to analyze the importance of each component and report the predictive and FTD scores. 

The supplementary materials consist of additional information on benchmarks, hyperparameters, training details, and visualizations for model evaluations. The implementation of GAT-GAN (pickled) can be found at: \href{https://github.com/srikrish010697/GAT-GAN}{https://github.com/srikrish010697/GAT-GAN}. 

\subsection{Datasets and baseline models}

The following publicly available time-series multivariate datasets are selected with different properties, including periodicity, level of noise, and spatial and temporal correlations. (1) \emph{Motor}: we consider the IEEE broken rotor bar dataset which is continuous-valued, periodic, and exhibits spatial correlations. We use the 3-phase voltage signals (Va, Vb, Vc) of a healthy induction motor to train the models \cite{fmnm-bn95-20},  (2) \emph{ECG}: we utilize the MIT-BIH Arrhythmia dataset (47 patients) of ECG recordings \cite{Moody_Arrhythmia_2001} and (3) \emph{Traffic}: we use the UCI PEMS-SF dataset consisting of hourly occupancy rates of lanes in San Francisco \cite{cuturi_fast_2011}. The dataset is continuous-valued and spatiotemporal in nature. The input dataset $\chi = [X_1,X_2,...,X_f]^T \in \Re^{K \times \tau \times F}$, where $\tau \in [16,64,128,256]$ and $F$ features. We perform min-max normalization for values to be within $[0,1]$ for all experiments in this paper. We compare GAT-GAN with different GAN architectures including TimeGAN \cite{yoon_time-series_2019}, SigWCGAN \cite{ni_conditional_2020}, RCGAN \cite{esteban_real-valued_2017} and GMMN \cite{li_generative_2015} based on Frechet Transformer distance and predictive score. 

\subsection{Frechet Transformer distance}

As shown in Table \ref{table1}, the performance results are outlined for four sequence lengths for GAT-GAN and its baselines. The metrics are run 10 times and the mean and standard deviation is reported. For all sequence lengths, the GAT-GAN model consistently generated high-quality synthetic data in comparison to the benchmarks by producing the lowest FTD scores. For instance, for traffic data, the GAT-generated data achieved $68\%, 78\%, 71\%, 64\%$ lower scores than the next best benchmark for sequence lengths $16, 64, 128, 256$ respectively, which is statistically significant. 

\begin{table}[h]
\centering
\caption{Frechet Transformer distance (lower is better) of GAT-GAN in comparison to benchmarks for multiple time-series datasets. All models are repeated 10 times and their mean, and standard deviation is reported. Among the benchmarks, GAT-GAN-generated samples achieved the lowest FTD scores.}
\label{table1}
\resizebox{\textwidth}{!}{%
\begin{tabular}{@{}c|c|llllc@{}}
\toprule
Dataset &
  Length &
  \multicolumn{1}{c}{TimeGAN} &
  \multicolumn{1}{c}{SigCWGAN} &
  \multicolumn{1}{c}{GMMN} &
  \multicolumn{1}{c}{RCGAN} &
  GAT-GAN \\ \midrule
\multirow{4}{*}{Motor} &
  16 &
  17.315+-0.079 &
  22.364+-0.39 &
  19.816+-0.644 &
  21.77+-0.82 &
  \textbf{10.908+-0.643} \\
 &
  64 &
  6.36+-0.012 &
  6.557+-0.201 &
  5.819+-0.627 &
  4.569+-0.64 &
  \textbf{1.35+-1.09} \\
 &
  128 &
  6.382+-0.164 &
  7.157+-0.093 &
  9.139+-0.993 &
  4.561+-0.69 &
  \textbf{1.187+-0.646} \\
 &
  256 &
  8.84+-0.601 &
  5.033+-0.382 &
  6.897+-0.59 &
  9.55+-0.709 &
  \textbf{4.038+-0.306} \\ \cmidrule(r){1-7}
\multirow{4}{*}{ECG} &
  16 &
  1.566+-0.96 &
  1.316+-0.682 &
  1.092+-0.886 &
  0.954+-0.348 &
  \textbf{0.527+-0.301} \\
 &
  64 &
  0.7+-0.486 &
  0.613+-0.364 &
  0.798+-0.572 &
  0.845+-0.373 &
  \textbf{0.42+-0.214} \\
 &
  128 &
  0.695+-0.397 &
  1.043+-0.677 &
  0.929+-0.454 &
  1.023+0.585 &
  \textbf{0.181+-0.126} \\
 &
  256 &
  0.683+-0.821 &
  0.568+-0.489 &
  0.669+-0.616 &
  0.714+-0.292 &
  \textbf{0.161+-0.118} \\ \cmidrule(r){1-7}
\multirow{4}{*}{Traffic} &
  16 &
  \multicolumn{1}{c}{10.516+-0.928} &
  \multicolumn{1}{c}{12.687+-0.333} &
  \multicolumn{1}{c}{13.107+-0.701} &
  \multicolumn{1}{c}{12.72+-0.751} &
  \textbf{3.383+-0.506} \\
 &
  64 &
  5.689+-0.442 &
  \multicolumn{1}{c}{7+-0.333} &
  \multicolumn{1}{c}{6.743+-0.256} &
  \multicolumn{1}{c}{6.021+-0.402} &
  \textbf{1.25+-0.667} \\
 &
  128 &
  6.495+-0.121 &
  \multicolumn{1}{c}{7.924+-0.361} &
  \multicolumn{1}{c}{4.901+-0.777} &
  \multicolumn{1}{c}{7.374+-0.82} &
  \textbf{1.406+-0.57} \\
 &
  256 &
  8.134+-0.759 &
  \multicolumn{1}{c}{8.034+-0.388} &
  \multicolumn{1}{c}{5.704+-0.2833} &
  \multicolumn{1}{c}{9.089+-0.57} &
  \textbf{2.055+-0.424} \\ \bottomrule
\end{tabular}%
}
\end{table}

\subsection{Predictive score: Evaluation on downstream forecasting}

A forecasting model was trained using synthetically generated data with a context length of $\tau - p$ to predict the next $p=8$ samples (prediction window). The prediction window is fixed at 8 to ensure that our LSTM model is not limited by its ability to model long-term dependencies. From Table \ref{table2}, we can observe that the GAT-GAN model has the least MAE (Mean absolute error) for medium-term (64) and long-term (128,256) sequence lengths based on a downstream forecasting task. For example, in the motor dataset, for a short sequence 128, with a context length = 120  and prediction window = 8, GAT-GAN generated samples achieve 0.127 which is $74\%$ lower than the next benchmark (RCGAN = 0.51). The explanation for this behavior is that GAT-GAN consists of a temporal graph attention block in the encoder which can extract long-term dependencies (unlike RNN-based models). Furthermore, we can observe that while the forecasting performance of benchmark algorithms degrades with increasing context lengths, GAT-GAN achieved nearly consistent scores across all context lengths. This is expected as GAT-GAN takes into account, both spatial and temporal dependencies with dynamic attention to preserve the underlying distribution of the input data and adaptively capture relationships between neighboring and distant nodes. However, note that for sequence length = 16, GAT-GAN produced competitive results with the baseline. 

\begin{table}[ht]
\centering
\caption{Comparison of downstream forecasting experiments based on predictive score (Mean Absolute error). All models are repeated 10 times and their mean, standard deviation is reported. Among the benchmarks, forecasting on GAT-GAN-generated samples had the lowest (lower is better) predictive scores for long-term sequences and competitive scores for short-term sequences.}
\label{table2}
\resizebox{\textwidth}{!}{%
\begin{tabular}{@{}c|c|lllll@{}}
\toprule
Dataset & Length & \multicolumn{1}{c}{TimeGAN} & \multicolumn{1}{c}{SigCWGAN} & \multicolumn{1}{c}{GMMN} & \multicolumn{1}{c}{RCGAN} & \multicolumn{1}{c}{GAT-GAN} \\ \midrule
\multirow{4}{*}{ECG}     & 16  & 0.061+-0.004  & \textbf{0.053+-0.0003} & 0.058+-0.007          & 0.058+-0.002  & 0.06+-0.001             \\
                         & 64  & 0.121+-0.001  & 0.148+-0.0001          & 0.149+-0.0001         & 0.151+-0.0001 & \textbf{0.049+-0.0003}  \\
                         & 128 & 0.152+-0.001  & 0.147+-0.0001          & 0.148+-0.0001         & 0.154+-0.0005 & \textbf{0.0478+-0.0002} \\
                         & 256 & 0.154+-0.002  & 0.167+-0.0001          & 0.156+-0.072          & 0.168+-0.003  & \textbf{0.047+-0.0001}  \\ \cmidrule(r){1-7}
\multirow{4}{*}{Traffic} & 16  & 0.027+-0.0001 & 0.034+-0.001           & \textbf{0.02+-0.0002} & 0.027+-0.0001 & 0.03+-0.0002            \\
                         & 64  & 0.141+-0.0004 & 0.107+-0.001           & 0.13+-0.0002          & 0.136+-0.0006 & \textbf{0.017+-0.0001}  \\
                         & 128 & 0.14+-0.019   & 0.118+-0.001           & 0.124+-0.001          & 0.149+-0.009  & \textbf{0.016+-0}       \\
                         & 256 & 0.134+-0.09   & 0.109+-0               & 0.18+-0.005           & 0.129+-0.005  & \textbf{0.004+-0.0001}  \\ \cmidrule(r){1-7}
\multirow{4}{*}{Motor}   & 16  & 0.354+-0.008  & 0.385+-0.01            & 0.339+-0.001          & 0.347+-0.007  & \textbf{0.161+-0.0001}  \\
                         & 64  & 0.157+-0.002  & 0.497+-0               & 0.14+-0.0008          & 0.147+-0.002  & \textbf{0.127+-0.0001}  \\
                         & 128 & 0.686+-0.021  & 0.741+-0.075           & 0.536+-0.037          & 0.51+0.061    & \textbf{0.135+-0}       \\
                         & 256 & 0.492+-0.02   & 0.712+-0.021           & 0.473+-0.006          & 0.493+-0      & \textbf{0.133+-0}       \\ \cmidrule(r){1-7}
\end{tabular}%
}
\end{table}

\subsection{Ablation study}

To analyze the importance of each component in GAT-GAN, we report the Frechet Transformer distance (FTD) score with the following modifications to GAT-GAN : (i) without the decoder (ii) without spatial-oriented graph attention (iii) without time-oriented graph attention, (iv) without Convolution in the encoder and (v) without reconstruction loss. We observe in Table \ref{table3} that all the five major components of the model play an important role in improving the quality of the generated samples. The spatial and temporal attention layers are vital components responsible for capturing robust representations, especially when data is quasi-periodic and exhibit high spatial correlations, such as in the motor dataset. The reconstruction-based elements (decoder and reconstruction loss) are crucial in finetuning the encoder while it is adversarially trained. Additionally, the usage of convolution feature maps before and after the graph attention layers has proven to be useful in improving generative performance across all datasets. 

\begin{table}[h]
\centering
\caption{The mean FTD scores with a standard deviation of the ablation study is shown by running the model 10 times for a fixed sequence length = 64}
\label{table3}
\resizebox{\textwidth}{!}{%
\begin{tabular}{@{}llll@{}}
\toprule
Method                       & Motor               & ECG                   & Traffic              \\ \midrule
GAT-GAN                      & \textbf{1.35+-1.09} & \textbf{0.42+-0.214}           & \textbf{1.25+-0.667} \\
w/o decoder                  & 1.441+-1.15         & 0.445+-0.204 & 1.896+-1             \\
w/o Spatial graph attention  & 1.535+-2.17         & 0.468+-0.384          & 1.557+-0.72          \\
w/o Temporal graph attention & 1.978+-1.229        & 0.434+-0.23           & 1.339+-0.875         \\
w/o convolution in encoder   & 2.082+-1.907        & 0.454+-0.228          & 1.679+-1.3           \\
w/o reconstruction loss      & 2.404+-2.197        & 0.429+-0.221          & 1.532+-1.131      
\\ \cmidrule(r){1-4}
\end{tabular}%
}
\end{table}

\subsection{Relationship between low FTD score and downstream forecasting performance}
From Table \ref{table1},\ref{table2}, we can observe that the FTD and predictive scores for GAT-GAN are lower with increasing sequence lengths. GAT-GAN has the lowest FTD scores, and outperforms almost all the benchmarks on the downstream forecasting experiment, except for short-term sequences (seq length = 16), the sigCWGAN and GMMN achieve slightly lower scores. We find that the Pearson's correlation coefficient between FTD scores and MAE for GAT-GAN (by sequence length) is $0.79$ averaged across all datasets. Hence, the model generating the lowest FTD score usually corresponds to the best model in the forecasting experiment. 

\section{Conclusion}

In this paper, we introduced GAT-GAN, a graph-attention-based GAN model capable of generating long-term realistic synthetic data by capturing spatial and temporal dependencies. We introduce a robust training procedure to adversarially train an autoregressive-based model, leveraging the contributions of reconstruction loss and the decoder. We proposed a metric known as Frechet Transformer distance to evaluate GAT-GAN in comparison to the state-of-the-art benchmarks. GAT-GAN demonstrated consistent and significant improvements over the benchmarks based on FTD scores. We also observed that the lowest FTD scores correspond to the best models in the downstream forecasting task. The routine use of GANs in time series modeling may be achieved through their ability to scale to longer sequences by capturing spatiotemporal features. Hence, in addition to assessing the fidelity and diversity of generated samples, future work can investigate how generalization in GANs can be evaluated to ensure generated samples are not merely \emph{copied} and data privacy is preserved.
\bibliographystyle{unsrt}
\bibliography{manuscript}

\end{document}